\documentclass[table]{article}

\usepackage{microtype}
\usepackage{graphicx}
\usepackage{booktabs} 

\usepackage{hyperref}

\usepackage[accepted]{icml2024}

\usepackage[utf8]{inputenc} 
\usepackage[T1]{fontenc}    
\usepackage{hyperref}       
\usepackage{url}            
\usepackage{booktabs}       
\usepackage{amsfonts}       
\usepackage{nicefrac}       
\usepackage{microtype}      
\usepackage{xcolor}
\usepackage{amsmath}
\usepackage{amssymb}
\usepackage{mathtools}
\usepackage{amsthm}
\usepackage{xspace}
\usepackage{subcaption}
\usepackage{comment}
\usepackage{wrapfig,lipsum,booktabs}
\usepackage{multicol}
\usepackage{multirow}

\newcommand{\X}{\mathcal{X}}

\newcommand{\I}{\mathcal{I}}
\newcommand{\Loss}{\mathcal{L}}
\newcommand{\T}{\mathcal{T}}
\newcommand{\D}{\mathcal{D}}
\newcommand{\NN}{\text{NN}}
\newcommand{\z}{\pmb{z}}
\newcommand{\x}{\pmb{x}}
\newcommand{\alg}{\textsc{SafeClip}\xspace}

\definecolor{GoodGreen}{rgb}{0.67, 0.88, 0.69} 
\definecolor{BadRed}{rgb}{0.98, 0.68, 0.82}   
\usepackage[capitalize,noabbrev]{cleveref}

\theoremstyle{plain}

\theoremstyle{definition}

\theoremstyle{remark}

\usepackage[textsize=tiny]{todonotes}

\icmltitlerunning{Pre-training CLIP against Targeted Data Poisoning and Backdoor Attacks}

\begin{document}

\twocolumn[
\icmltitle{Better Safe than Sorry: Pre-training CLIP against \linebreak Targeted Data Poisoning and Backdoor Attacks}



\icmlsetsymbol{equal}{*}

\begin{icmlauthorlist}
\icmlauthor{Wenhan Yang}{yyy}
\icmlauthor{Jingdong Gao}{yyy}
\icmlauthor{Baharan Mirzasoleiman}{yyy}
\end{icmlauthorlist}

\icmlaffiliation{yyy}{Department of Computer Science, University of California, Los Angeles, Los Angeles, CA, 90024}

\icmlcorrespondingauthor{Wenhan Yang}{hangeryang18@g.ucla.edu}

\icmlkeywords{Machine Learning, ICML}

\vskip 0.3in
]



\printAffiliationsAndNotice{\icmlEqualContribution} 

\begin{abstract}
Contrastive Language-Image Pre-training (CLIP) 
on large image-caption datasets has
achieved remarkable success in zero-shot classification and enabled transferability to new domains. However, CLIP is extremely more vulnerable to targeted data poisoning and backdoor attacks, compared to supervised learning. Perhaps surprisingly, poisoning 0.0001\% of CLIP pre-training data is enough to make targeted data poisoning attacks successful. This is four orders of magnitude smaller than what is required to poison supervised models. Despite this vulnerability, existing methods are very limited in defending CLIP models during pre-training.
In this work, we propose a strong defense, \alg, to safely pre-train CLIP against targeted data poisoning and backdoor attacks. \alg  warms up the model by applying unimodal 
contrastive learning (CL) on image and text modalities separately. Then, it divides the data into safe and risky sets, by applying a Gaussian Mixture Model to the cosine similarity of image-caption pair representations. \alg pre-trains the model by applying the CLIP loss to the safe set and applying unimodal CL to image and text modalities of the risky set separately.
By gradually 
increasing the size of the safe set during pre-training, 
\alg effectively breaks targeted data poisoning and backdoor attacks without harming the CLIP performance. 
Our extensive experiments on CC3M, Visual Genome and MSCOCO demonstrate that \alg significantly reduces 
the success rate of targeted data poisoning attacks from 93.75\% to 0\% and that of various backdoor attacks from up to 
100\% to 0\%, without harming CLIP's performance\footnote{Code can be found at \url{https://github.com/BigML-CS-UCLA/SafeCLIP}}.

\end{abstract}
\section{Introduction}
Pre-training large vision-language models on extensive image-caption data crawled from the internet has achieved remarkable success in zero-shot classification and robustness to distribution shift. CLIP learns image and text representations in a shared space by maximizing the agreement between the paired image-text representations, and minimizing the agreement between the unpaired ones. This alleviates the need for high-quality annotations and allows scaling up the pre-training data to millions \citep{radford2021learning} and billions of examples \citep{jia2021scaling}.
Despite its superior performance, CLIP is extremely vulnerable to targeted data poisoning and backdoor attacks, where an adversary injects a subset of malicious examples in the training data to change the prediction of particular examples at test time. Perhaps surprisingly, poisoning only 0.0001\% and 0.01\% of the pre-training data is enough to make targeted data poisoning and backdoor attacks successful, respectively \textcolor{blue}{\citep{carlini2023poisoning,carlini2021poisoning}}. Considering that the large pre-training data of CLIP is often crawled from the internet, such attacks are very easy to perform in practice. \looseness=-1

Despite this vulnerability, protecting CLIP against targeted data poisoning and backdoor attacks during pre-training has remained largely unaddressed. The only recently proposed method, RoCLIP, aims to disassociate the poisoned image-caption pairs during pre-training {by pairing each image representation with 
its most similar caption representation in a random caption pool \citep{yang2023robust}. However, RoCLIP can suffer significant performance drop in downstream performance, limiting its real-world application.} Two other methods proposed to clean an already \textit{poisoned pre-trained} CLIP, by fine-tuning on a \textit{clean} data of the same scale as pre-training \citep{yang2023data}, or fine-tuning on a \textit{clean} subset of pre-training data using contrastive learning on image and text modalities \citep{bansal2023cleanclip}. The first method is clearly not applicable to pre-training, and the second one even increases the attack success rate if applied 
during 
pre-training on poisoned data, as confirmed in \cite{yang2023robust}. 

Protecting CLIP against targeted data poisoning and backdoor attacks during pre-training is indeed very challenging. This is because training only once on the poisoned pairs can make the attack successful. In contrast, in the supervised setting the model should be trained on the poisoned data for several epochs before the attack succeeds \citep{biggio2012poisoning, turner2019label}. Thus, to protect CLIP during pre-training, it is cruical to entirely exclude the poisoned examples from the pre-training pipeline. \looseness=-1

{In this work, we propose an effective defense, \alg, against {strong} targeted data poisoning and backdoor attacks during pre-training CLIP, {without compromising its performance.}}
\alg warms up the model by applying separate unimodal contrastive losses to image and caption modalities to reduce the initial similarity of poisoned image-caption representations. 
Then, it applies the CLIP loss once to all pairs with a {low learning rate} to initially associate the image-caption representations, while maintaining a low similarity for poisoned pairs.
{Subsequently, \alg employs a Gaussian Mixture Model (GMM) on cosine similarity of image-caption representations to divide the examples into {safe} and {risky} sets.} \alg pre-trains the model using the CLIP loss on the {safe} set and unimodal contrastive losses on image and caption modalities of the {risky} set. Throughout training, \alg updates and expands the {safe} set. In doing so, it effectively prevents the poisoned image-caption pairs to be associated and successfully breaks the attack. At the same time, it maintains model performance with the increasing training data size. 

We conduct extensive experiments on three image-caption datasets with different sizes and data distributions, namely Conceptual Captions 3M (CC3M) \citep{sharma2018conceptual}, Visual Genome (VG) \citep{krishna2017visual}, and MSCOCO \cite{lin2014microsoft}, that are poisoned with various targeted data poisoning and backdoor attacks. 
We show that \alg successfully defends CLIP against targeted data poisoning and backdoor attacks during pre-training, reducing success rate of targeted poisoning attacks from 93.75\% to 0\%, and backdoor attacks from up to 100\% to 0\%, without compromising CLIP's zero-shot and linear prob performance.\looseness=-1

\vspace{-2mm}
\section{Related Work} 
\vspace{-1mm}
\textbf{Unimodal Contrastive Learning (CL)}
Unimodal contrastive learning is among the most successful methods for representation learning \citep{chen2020simple, caron2020unsupervised, chen2021exploring}. CL maximizes the agreement between different augmented views of the same example (positive pairs) while minimizing it for different examples (negative pairs). A recent body of work aimed to further improve the performance of CL, by improving the consistency of the representations via momentum encode \citep{he2020momentum},  eliminating the need for negative pairs  \citep{grill2020bootstrap}, or removing redundancy between components of the representation vectors \citep{zbontar2021barlow}.
Most relevant to our work is NNCLR, which enriches the learned representations by keeping a memory bank of augmented representations and using each example's nearest neighbor in it as its positive pair \citep{dwibedi2021little}. \looseness=-1

\textbf{Contrastive Language-Image pre-training (CLIP)} 
Large vision-language models like CLIP \citep{radford2021learning} and ALIGN \citep{jia2021scaling} achieved a remarkable success by contrastive pre-training on 400M and 1B image-caption pairs crawled from the web. Recent work tried to improve the data efficiency and performance of CLIP. Specifically, DeCLIP \citep{li2021supervision} uses SimSiam \citep{chen2021exploring} and Masked Language Modeling \citep{devlin2018bert} to match the augmented views of the image representations and the augmented views of the text representations, to improve the data efficiency of CLIP. CyCLIP \citep{goel2022cyclip} emphasizes the importance of in-modal consistency and cross-modal consistency between text and image modality. SLIP \citep{mu2022slip} improves the performance by including unimodal contrastive learning on images using SimCLR, which maximizes the agreement between different views of the same augmented image while minimizing agreement between augmented views of different images. 

\textbf{Targeted Data Poisoning and Backdoor Attacks on CLIP }
CLIP is highly susceptible to various types of targeted data poisoning and backdoor attacks \citep{carlini2021poisoning, yang2023data}. Targeted data poisoning attacks (TDPA) aim to deceive the model into misclassifying a specific test example by modifying the captions of a small subset of the training data. Backdoor attacks (BA) involve embedding a backdoor trigger into a small subset of examples in the training data, with the goal of causing the model to misclassify any test images with the same trigger. A backdoor trigger can be either visible, like a distinguishable patch, or invisible, like patterned noise points or patterned image deformation \citep{chen2017targeted, gu2017badnets, nguyen2021wanet}. 
Adding trigger to only $0.01\%$ of the pre-training data can cause the model to misclassify the backdoored examples. TDPA is even more effective, requiring only 0.0001\% of the data to be poisoned \citep{carlini2021poisoning}. \looseness=-1

\textbf{Protecting CLIP against Targeted Data Poisoning and Backdoor Attacks}
Despite the vulnerability of CLIP to TDPA and BA, existing defense methods are very limited. RoCLIP \citep{yang2023robust} is the only proposed defense for protecting CLIP during pre-training. RoCLIP first augments image-caption pairs using techniques such as random cropping and color jittering. {Subsequently, it matches each image with its nearest-neighbor caption in a pool of random captions.} 
The caption
representation pool is 
updated at the end of every epoch.  RoCLIP, however, may lead to a significant performance drop when defending a higher amount of poisons.

Two recent works proposed data cleansing for fine-tuning CLIP, or cleaning a poisoned pre-trained CLIP during fine-tuning. \cite{yang2023data} proposed dropping examples that have a low image-caption similarity based on a clean pre-trained CLIP, to cleanse the fine-tuning data. This method requires a clean pre-trained model, and a proper threshold to filter the poisons without discarding a large amount of clean data. This threshold varies for different attack types and is difficult to pre-compute.
To clean a poisoned CLIP with TDPA, \cite{yang2023data} proposed fine-tuning on a clean dataset of the same size as the pre-training data. Moreover, to clean a poisoned CLIP with BA, \cite{bansal2023cleanclip} proposed CleanCLIP, which fine-tunes the model on a \textit{clean} subset of the pre-training data with CLIP loss and CL loss on image and text modalities. The first method is clearly not applicable to pre-training and the second one, as shown in \cite{yang2023robust}, can increase the attack success rate when applied to the poisoned data. This is because CL cluster the backdoored images and their cpations, and the CLIP loss can even better associate the backdoored images with the poisoned captions.

In this work, we propose the first effective defense for protecting CLIP against strong TDPA ($0.05\%$) and BA ($0.05\%-0.15\%$) during pre-training, without compromising the model's performance. 
\section{Preliminary}
\begin{figure} 
    \centering
        \centering
        \includegraphics[width=1\columnwidth]{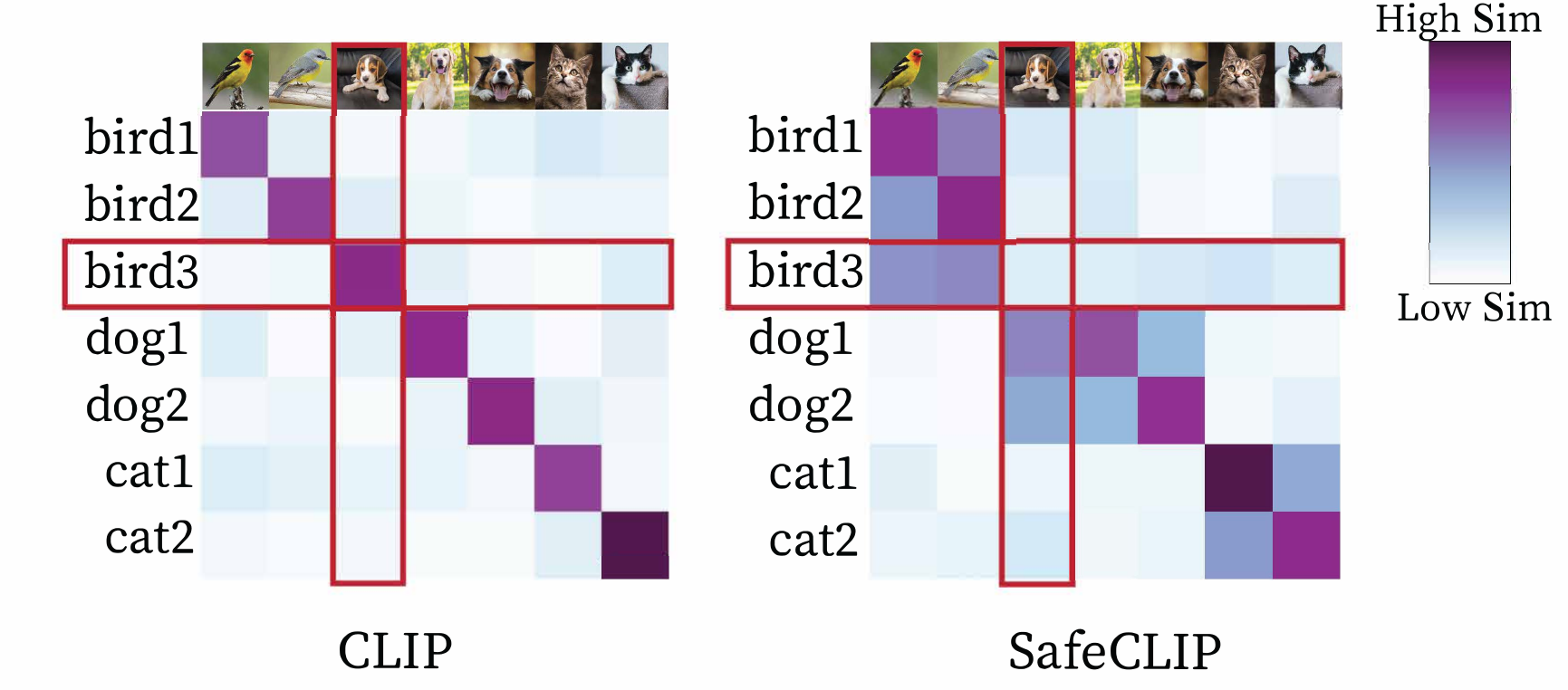} %
    \vspace{-5mm}
    \caption{Cosine similarities between image-caption representations. While CLIP directly associate the poisoned image-caption pairs, \alg clusters the images and captions in the same category and pushes away poisoned pairs.\looseness=-1}
\label{fig:matrix}
    \vspace{-5mm}
\end{figure}
\subsection{Contrastive Language-Image Pre-training (CLIP)}\label{sec:pre_clip}
Consider a dataset $\D = \{(\x_i^\I, \x_i^\T)\}_{i=1}^n$ of $n$ image-captions pairs, where $\x_i^\I$ and $\x_i^\T$ are the image and caption of the $i^{th}$ pair. The CLIP architecture consists of an image encoder $f_I\!:\!\I \rightarrow \mathbb{R}^d$ and a text encoder $f_T\!:\!\T\rightarrow \mathbb{R}^d$ to encode images and captions. The encoded representations are projected into the same space and are normalized to have unit $\ell_2$-norm. We denote the resulting image and text representations by $\z_i^\I, \z_i^\T$. To create the multi-modal interaction, the InfoNCE loss is applied to pull the projected representations of every image-caption pair together while pushing apart the projected representations of unpaied images and captions in the same mini-batch. Formally, for a mini-batch of $N$ pairs, the CLIP loss is defined as: 
\begin{equation}\label{eq:clip}
\begin{split}
    \mathcal{L}_{\text{CLIP}} = &-\frac{1}{2N} \sum_{j=1}^N \log \left [ \frac{\exp\left(\left<\z_j^\I,\z_j^\T\right>/\tau\right)}{\sum_{k=1}^N \exp\left(\left<\z_j^\I,\z_k^\T\right>/\tau\right) } \right] \\
    &-\frac{1}{2N} \sum_{k=1}^N \log \left [ \frac{\exp\left(\left<\z_k^\I,\z_k^\T\right>/\tau\right)}{\sum_{j=1}^N \exp\left(\left<\z_j^\I,\z_k^\T\right>/\tau\right) } \right],
\end{split}
\end{equation}
where $\tau$ is a trainable temperature parameter, and $\left<.,.\right>$ is the inner product between two representations. 
The performances of CLIP is evaluated with zero-shot or linear-probe, as we discuss next. \looseness=-1

\textbf{Zero-shot classification.} 
Zero-shot classification 
assess the generalizability and transferability of the model to unseen tasks. It transforms the downstream labels into natural language captions using the provided engineered prompt templates, such as "\texttt{A photo of a \{label\}}" \citep{radford2021learning}. Then, it calculates the cosine similarity between the representations of a given image and each prompt, and predicts the label with the highest image-prompt similarity.\looseness=-1

\textbf{Linear probe classification.} 
Linear probe classification refers to evaluating the extracted representations from the pre-trained image encoder for training a linear classifier on the downstream labeled data. \looseness=-1

\subsection{Targeted Data Poisoning and Backdoor Attacks}\label{sec:pre_attack}
Targeted data poisoning and backdoor attacks poison CLIP by injecting a set of poisoned image-caption pairs to the pre-training data.
Let $\D_p = \{(\x_i^\I, \x_c^\T)| \x_i^\I\in\I_t, \x_c^\T \in \T_{adv}\}$ be the injected poisoned pairs, where $\I_t$ is the poisoned image(s) and $\T_{adv}$ is the set of adversarial caption related to the adversarial label $y_{adv}$. 
To construct the poisoned caption set, one can search the training dataset for all captions that contain the adversarial label and use these captions as the adversarial captions. Another approach is to use CLIP's set of 80 different prompt-engineered text descriptions \citep{radford2021learning} to construct captions for the adversairal label, and then either use a subset of them or repeat them as necessary. In our work, we construct $\T_{adv}$ from the training dataset, which is consistent with the construction methods used in \citep{carlini2021poisoning, yang2023data,yang2023robust,bansal2023cleanclip}. \looseness=-1

\textbf{Targeted data poisoning attacks} aim to misclassify a particular test example, $\x_i^\I$, as $y_{adv}$. Hence,  
$D_p=\{(\x_i^\I,\x_c^\T)|\x_c^T\in\T_{adv}\}$. \looseness=-1


\textbf{Backdoor attacks} introduce a trigger patch to a set of poisoned images. The goal is to misclassify any test examples with the trigger patch, $\x_i^\I \oplus \text{patch}$, as $y_{adv}$. Hence, $D_p=\{(\x_i^\I\oplus \text{patch},x_c^\T)| \x_i^\I\in\I, \x_c^T\in\T_{adv}\}$.
In contrast to targeted data poisoning attacks 
which target a particular test example, 
backdoor attacks inject \textit{random} images with the backdoor trigger, paired with the adversarial captions.

\textbf{Adversary Objective} The primary objective of the adversary is to manipulate the output representations of CLIP, such that certain images are misclassified into adversarial categories instead of their true categories, while the other images are classified correctly.\looseness=-1

\textbf{Adversary Capabilities} We assume that the adversary has limited control over the pre-training data, and can inject a small number of poisoned examples ($\le 0.05\%$ of the dataset size for TDPA and $\le 0.15\%$ of the dataset size for BA) into the training dataset. Adversary also has the knowledge of the model structure, the training algorithm, and the hyperparameter used by their victim, but they cannot modify the training process directly.\looseness=-1

\vspace{-2mm}
\section{Method}\vspace{-1mm}

\textbf{Motivation} Targeted data poisoning and backdoor attacks can succeed extremely fast when pre-training CLIP models. For example, {when pre-training on a dataset with 0.01\% poison rate, as shown in Appendix, Fig. \ref{fig:noseparation}, the poisoned pairs become inseparable from the clean pairs after 1 pre-training epochs.} Thus, to prevent the model from being poisoned, it is essential to filter out the majority of poisoned pairs \textit{before} the pre-training starts, and keep them out \textit{throughout} the pre-training. 
If the model avoids training on or is exposed to only a limited amount of the poisoned data, the representations of poisoned images and captions do not get close during pre-training, and the attack fails. 

\textbf{Main Idea} To achieve this, \alg warms up the model with a few unimodal CL epochs on image and text modalities separately. In doing so, it clusters similar images and texts, and thus pushes away poisoned images from their adversarial captions that belong to another category. Subsequently, \alg 
applies the CLIP loss once to all examples with a very \textit{small learning rate} to associate large clusters of similar image-caption pairs. As a result, as poisoned images and their captions are pushed apart during unimodal CL warmup, their cosine similarity remains small. 
This warmup helps separate poisoned pairs from clean pairs. \alg then separates image-caption pairs into a \textit{safe} set containing examples with very high cosine similarity between their image-caption representations, and a \textit{risky} set otherwise.
Subsequently, it pre-trains the model by applying the CLIP loss to data in the safe set and unimodal CL loss to data in the risky set. Then, \alg gradually increases the size of the safe set. This method maintains a low poison ratio in the safe set, effectively defending against strong attacks while boosting the downstream performance. In summary, to prevent the model from being poisoned, \alg consists of three steps:
(1) A few epochs of unimodal CL warmup;
(2) Applying CLIP loss with very small learning rate to all examples;
(3) Pre-training 
with CLIP loss on safe set with high cosine similarity, and unimodal loss on the risky set, while gradually increasing the size of the safe set. 
The effect of \alg on image and text encoders is shown in Fig \ref{fig:matrix}. CLIP directly aligns the paired image-caption representations, and is thus prone to being poisoned. On the other hand, \alg only clusters images and captions in the same category. In doing so, it reduces the similarity of poisoned image-caption representations, which allows \alg to successfully defend strong poisoning and backdoor attacks.

Next, we will discuss each step in more details.
\begin{figure}
    \centering
    \includegraphics[width=0.65\columnwidth]{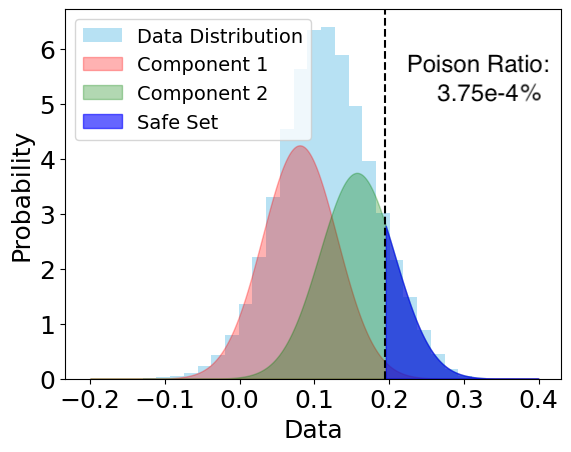} 
    \vspace{-3mm}
    \caption{\alg fits a two-components Gaussian Mixture Model (GMM) to the post-warmup cosine similarity, selecting the safe set based on the chosen threshold $t$. This approach reduces the poison rate to as low as $3.75e^{-4}\%$.}\label{fig:gmm}
    \vspace{-5mm}
\end{figure}

\subsection{Unimodal 
CL Warmup: Pushing Adversarial Captions away from Poisoned Images
}\label{sec:warmup}
\alg applies unimodal CL to image and text modalities separately. In doing so, it clusters similar images and captions while keeping poisoned images apart from their adversarial caption. 
{Effectively, during unimodal CL warmup, poisoned images and adversarial captions, belonging to different categories, cluster with examples in their own category and move away from each other in the representation space. For example, to poison an image of `cat' with a `plane' caption, the image needs to move closer to the plane text cluster and away from the cat image cluster in the representation space. The closer the image is to its true cat representation cluster at the beginning of training, the more challenging it becomes to poison the image. Same argument applies to captions.} 
As unimodal CL does not match poisoned images with captions, it does not risk poisoning the models.
Only $r\!\!=\!\!5$ epochs of unimodal CL is sufficient on various various datasets and attack types, as we will confirm in Sec. \ref{sec:abl_warmup}. \looseness=-1

\textbf{Nearest-Neighbors}
When the poison rate is high, poisoned images, which are either identical images (TDPA) or images sharing the backdoor patch (BA) cluster tightly together in the representation space. 
To avoid this and enrich the representation quality, we incorporate a nearest neighbor (NN) pool in our unimodal CL training for finding positive pairs  \citep{dwibedi2021little}.
Instead of matching augmented views of the same image or caption, we match each representation with its NN 
in a random 
pool of image representations. The pool is initialized with random example representations and is updated with current mini-batch representations, displacing the oldest in the pool. By introducing more diverse positive pairs, \alg prevents clustering of poisoned images and adversarial captions, and can separate the poisoned pairs more effectively, as we will empirically confirm in Sec. \ref{sec:abl_warmup}. 
The unimodal CL loss is defined as:
\begin{align}\label{eq:unimodal_nnclip}
    &\mathcal{L}_{\text{unimodal}}
    =-\log \! \frac{\exp\left(\left<{\NN(\z_i, \mathcal{P}), \z_i^+} \right>/\tau\right)}{\sum_{k=1}^N \exp\left(\left<\NN(\z_i, \mathcal{P}),\z_k^+\right>/\tau\right) } 
\end{align}
where $\z_i$ is the output image/text representation and $\z_i^+$ is the augmented view of the image/text representation, and $\NN(\z_i, \mathcal{P})$ is the NN operator defined as:
\begin{align}\label{eq:unimodal_nn}
NN(\z_i) = \text{argmin}_{\pmb{p} \in \mathcal{P}}\|\z_i-\pmb{p}\|_{2}.\vspace{-2mm}
\end{align}
%
\subsection{\hspace{-.5mm}Separating Safe \!\&\! Risky (Potentially Poisoned) Data}
\begin{figure}
    \centering
    \begin{subfigure}{0.49\columnwidth}
        \centering
        \includegraphics[width=1\textwidth]{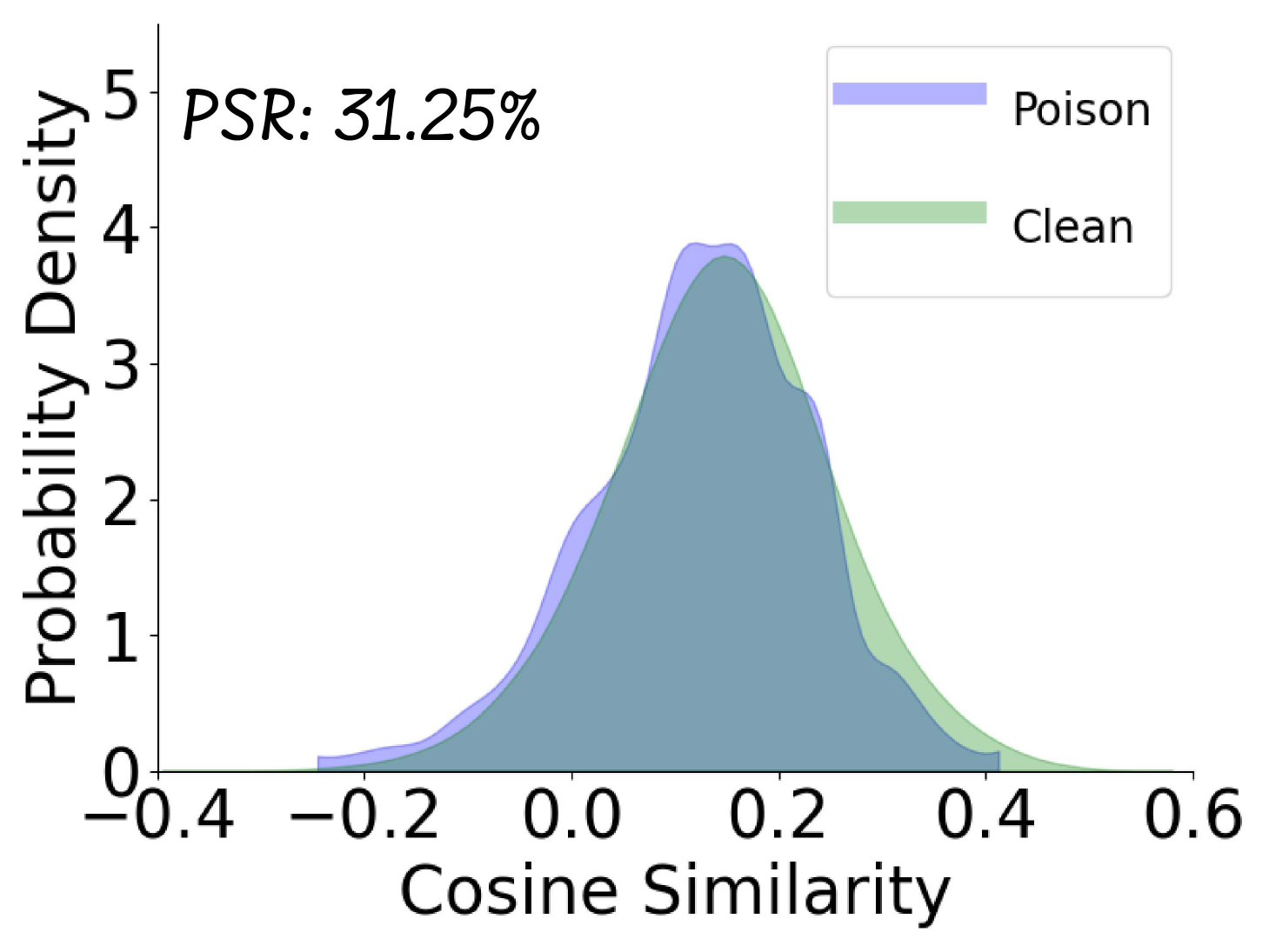} 
    \caption{CLIP}\label{fig:noseparation}
    \end{subfigure}\hfill
    \begin{subfigure}{0.49\columnwidth}
        \centering
        \includegraphics[width=1\textwidth]{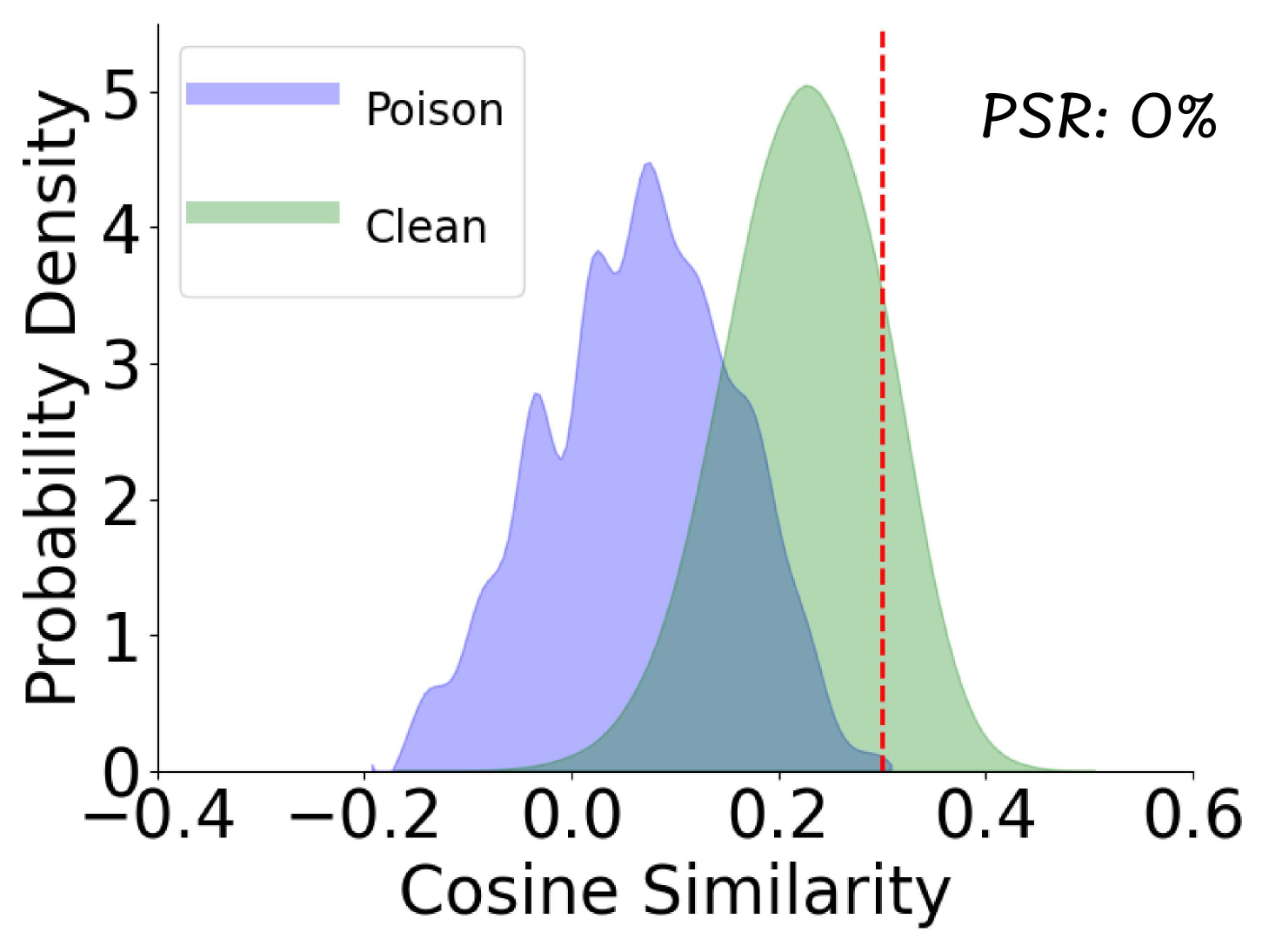} 
        \caption{\alg}\label{fig:separation}
    \end{subfigure}\hfill
    \hfill
    \caption{Distribution of Image-Caption Cosine Similarities After 1 epoch of Pre-Training with (a) CLIP and (b) \alg. While the poisoned pairs become indistinguishable from the clean pairs in CLIP, the warm-up helps \alg separate the clean data pairs from the poisoned data pairs. For clearer visualization, the distributions of poisoned and clean pairs are normalized.}\label{fig:warmup_dist}
\end{figure}

While unimodal CL 
clusters similar images and captions in their respective representation spaces, the image-caption pairs often remain relatively distant from each other. Thus, to effectively associate these image-caption representations and distinguish the potentially poisoned pairs, 
we apply the CLIP loss with a \textit{very low learning rate} once to all image-caption pairs.
In Sec.\ref{sec:abl_warmup}, we will 
confirm that lowering learning rate of CLIP by a factor of $0.01$ minimally associates the image-caption pairs without poisoning the model, 
across various datasets and attack types. 
As shown in Fig \ref{fig:separation}, the warmup results in a significant separation between poisoned and clean pairs. 

Subsequently, we calculate the cosine similarities of all pairs of image-caption representations and divide examples into a safe and a risky sets based on their cosine similarities. 
To do so,
we fit a two-component Gaussian Mixture Model (GMM) to the cosine similarities using the Expectation-Maximization (EM) algorithm \cite{permuter2006study}. For each image-caption pair $i$, we calculate 
the probability $p_i$ of its image-caption cosine similarity to be in
the Gaussian component with \textit{larger} mean, 
containing pairs with \textit{highest} cosine similarity.
We put pairs with a very high $p_i$, i.e., $p_i> t= 0.9$ into the \textit{safe} set, and put the remaining pairs in the \textit{risky} set. In our experiments (\textit{c.f.} Sec. \ref{sec:abl_warmup}), we show that 
threshold of $0.9$ is effective across different datasets and attack types. Fig. \ref{fig:gmm} shows how GMM successfully separates the safe and risky sets. 
By selecting only the data pairs with high confidence, \alg decreases the poison rate in the safe set (from initial $0.05\%$) to as low as $0.000375\%$.

\subsection{Applying CLIP to Safe and CL to Risky Data}
\alg pre-trains the model by applying the CLIP loss only to the safe data, to match their image-caption pairs. Meanwhile, rather than discarding the risky data, it continues to train on their images and captions separately using unimodal CL losses. This further helps separating clean and poisoned pairs, as discussed in the previous section. 
However, two concerns still remain:
(1) Some poisoned pairs may still be in the safe set;
(2) Model's performance may suffer as the CLIP loss is not applied to majority of examples. \looseness=-1

To address these concerns: (1) We apply data augmentation to the examples in the safe set used in the CLIP loss. 
Data augmentation has two advantages: Firstly, it can significantly strengthen defenses against various attacks \citep{yang2023robust}. Secondly, it improves the model's performance \cite{li2021supervision}. We use the SimCLR image augmentation method including random image cropping, horizontal flipping, color jittering, grayscale conversion, and blurring \citep{chen2020simple}. For text modality, we used the same Easy Data Augmentation proposed in \citep{wei2019eda}, which applies simple text token transformation like synonym replacement and random delete.
(2) Moreover, at the end of each epoch, we evaluate the cosine similarity of all examples. Then, we update the safe set and increase its size by $s=1\%$. 
Larger $s\%$ speeds up training, while exposing increase the risk of being poisoned. 
We empirically confirm that this conservative choice of $s=1\%$ is safe across various datasets and attacks.

With the above update strategy, even when few poisoned pairs enter the safe set, \alg can filter them out in the next epoch.
At the same time, more training on clean data with CLIP loss and on risky data with unimodal CL loss allows the model to learn better representations and better identify and discard the poisoned pairs during pre-training. 
Additionally, since we progressively increase the proportion of safe data during training, by the end of the training, the majority of the data will be part of the safe data and will be trained on with CLIP loss, 
thereby resolving the performance issue. 
The loss of the mixed training is defined as: \looseness=-1
\begin{align}\label{eq:safe_loss}
&\mathcal{L}_{\text{\alg}}(\D)=\mathcal{L}_{\text{unimodal}}(\D_\text{risky})+\mathcal{L}_{\text{CLIP}}(\D_\text{safe\_aug}).
\end{align} Note that, during mixed training, we still apply nearest-neighbors for unimodal CL. 

\alg's pseudocode is illustrated in Appendix, Alg. \ref{alg:alg}. 

\begin{table*}[ht!]
\caption{Effectiveness of \alg in defending against various adversarial attacks, measured by Attack Success Rate (ASR). 
\alg achieves a strong defense across datasets and attacks, outperforming RoCLIP by \textbf{37.5\%} on Visual Genome in defending against Targeted Data Poisoning Attacks (TDPAs) and by \textbf{4.6\%} in defending against Blended Backdoor Attacks (BA). Table \ref{tab: Classification Utility} shows that \alg maintains the performance of CLIP while RoCLIP drops it by \textbf{10\%}.}
\vspace{-2mm}
\begin{small}
\begin{tabular}{@{}lccccc|ccccc@{}} 
\toprule
\textbf{Dataset} & \multicolumn{5}{c}{\textbf{CC1M}} & \multicolumn{5}{c}{\textbf{CC3M}} \\
\cmidrule(lr){2-6} \cmidrule(l){7-11}
Attacks & TDPA & BadNet & Label Consis & Blended & WaNet & TDPA & BadNet & Label Consis & Blended & WaNet \\
\midrule
CLIP            & 93.75\% & 100\% & 71.0\% & 99.3\% & 96.3\% & 93.75\% & 100\% & 58.3\% & 100\% & 96\% \\
RoCLIP          & \textbf{0\%} & \textbf{0\%} & \textbf{0\%} & \textbf{0\% }& \textbf{0\%} & \textbf{0\%} & \textbf{0\%} & \textbf{0\%} & \textbf{0\%} & \textbf{0\%} \\
\textbf{\alg}        & \textbf{0\%} & \textbf{0\%} & \textbf{0\%} & \textbf{0\%} & \textbf{0\%} & \textbf{0\%} & \textbf{0\%} & \textbf{0\%} &  0.3\% &  \textbf{0\%} \\

\midrule
\textbf{Dataset} & \multicolumn{5}{c}{\textbf{MSCOCO}} & \multicolumn{5}{c}{\textbf{Visual Genome}} \\
\cmidrule(lr){2-6} \cmidrule(l){7-11}
Attacks & TDPA & BadNet & Label Consis & Blended & WaNet & TDPA & BadNet & Label Consis & Blended & WaNet \\
\midrule
CLIP            & 62.5\% & 31.0\% & 71.6\% & 95.3\% & 7.6\% & 62.5\% & 1.3\% & 28.6\% & 90.3\% & 18.6\% \\
RoCLIP          & \textbf{0\%} & \textbf{0\% }& \textbf{0\%} & 2.6\% & \textbf{0\%} & 37.5\% & \textbf{0\%} & \textbf{0\%} & 9.6\% &\textbf{ 0\%} \\
\textbf{\alg}         & \textbf{0\%} & \textbf{0\%} & \textbf{0\%} & \textbf{2\%} & \textbf{0\%} & \textbf{0\%} & \textbf{0\%} & \textbf{0\%} & \textbf{5\%} & \textbf{0\%} \\
\midrule
\bottomrule
\end{tabular}
\label{tab:defense}
\end{small}
\end{table*}
\setlength{\belowcaptionskip}{-2pt}

\begin{table*}[ht!]
\caption{Downstream linear probe and zero-shot (top 1) accuracy of pre-training on CC3M. The highest performance is bold and the lowest underscored. The last column highlights the average improvement over CLIP across 10 datasets. \alg, on average, achieves similar downstream performance to CLIP, while RoCLIP experiences a performance loss of nearly \textbf{10\%} in both linear probe and zero-shot evaluations.}
\vspace{-2mm}
\vspace{-2mm}
\label{tab: Classification Utility}
\begin{center}
\begin{small}
\begin{tabular}{p{1.1cm}p{0.9cm}p{0.5cm}cccccccccc}
\toprule
Method & Task & \rotatebox[origin=c]{90}{F102} & \rotatebox[origin=c]{90}{Fd101} & \rotatebox[origin=c]{90}{I1K} & \rotatebox[origin=c]{90}{Pet} & \rotatebox[origin=c]{90}{Cars} & \rotatebox[origin=c]{90}{Cal101}  & \rotatebox[origin=c]{90}{C10} & \rotatebox[origin=c]{90}{C100} & \rotatebox[origin=c]{90}{DTD} & \rotatebox[origin=c]{90}{Air.} &\rotatebox[origin=c]{45}{\textbf{Alg-CLIP}}\\
\midrule
 & 0-shot & \underline{16.7}  & \textbf{13.0} & \textbf{19.4} & \textbf{3.3}  &\textbf{ 1.2}  & \underline{50.8}   & \underline{48.2}  & \underline{18.9} & \underline{3.7} & 1.0& -\\
CLIP & lin-prb & \textbf{100}  & \textbf{54.8} & \underline{33.2}  & \textbf{58.8} & \underline{18.8} & \underline{80.2}  & \underline{77.8}  & \textbf{54.7}  & \underline{58.4}  & \textbf{28.5} & -\\
\midrule
 & 0-shot & 6.8  & 5.5 & 5.8 & \underline{2.6}  & 0.6  & 21.9   & 24.2  & 6.3 &  \textbf{4.6} & \textbf{1.1} & \textbf{-9.7}\\
RoCLIP & lin-prb & 91.2  & 47.9 & 21.8 & 47.8  & 17.4  & 66.5   & 67.1  & 45.9 & 53.5 & 23.3 & \textbf{-8.3}\\
\midrule
 & 0-shot & \textbf{17.5}  & \underline{11.1} & \underline{18.2} & 1.5 & \underline{0.9} &  \textbf{54.4}  & \textbf{54.7} & \textbf{22.6} & 3.6 & \textbf{1.1} & \textbf{+0.9}\\
\textbf{\alg} & lin-prb &\underline{99.8}  &  \underline{53.3} & \textbf{34.3} & \underline{58.1} & \textbf{21.3}  & \textbf{81.1} & \textbf{78.3}  & \underline{54.2}  & \textbf{62.9}  & \underline{26.9} & \textbf{+0.5}   \\
\midrule
\bottomrule
\end{tabular}
\end{small}
\end{center}
\end{table*}
\setlength{\belowcaptionskip}{-3pt}

\vspace{-3mm}
\section{Experiments}
\vspace{-2mm}
In this section, we evaluate the effectiveness of \alg against strong TDPA and BAs. We first introduce the experimental setup, and then present our main results. We finish by an ablation study on different components of \alg. \looseness=-1

\textbf{Pre-training Data}
To cover a wide range of dataset distributions, we consider three datasets with various distributions and sizes, namely Conceptual Captions 3M (CC3M) \citep{sharma2018conceptual}, Visual Genome (VG) \citep{krishna2017visual}, and MSCOCO \cite{lin2014microsoft}. Additionally, following \cite{yang2023robust}, we randomly sample 1M image-caption pairs from CC3M (termed CC1M) to demonstrate \alg's defense capabilities in datasets of varying sizes. The details of each dataset are listed in Appendix \ref{sec:dataset}.
We consistently employ a single set of hyperparameters, {i.e., $s\!=\!1\%$, $r\!=\!5$, $lr_\text{low}\!=\!5e^{-6}, t=0.9$}, across all our experiments. This demonstrates that \alg can provide effective defense against different types of attacks, across various datasets with different sizes and distributions.

\textbf{Setup}
We use open-source implementation of CLIP as our base model. Similar to the setup in \citep{radford2021learning}, we utilize a ResNet-50 as the image encoder and a transformer as the text encoder. In each experiment, except RoCLIP, all models are trained from scratch for 48 epochs.
For RoCLIP, we set the matching frequency to 2, as required for defense against a high poison rate of 0.05\%, and train for 24 epochs as recommended, as more training significantly increases the attack success rates \cite{yang2023robust}.\looseness=-1

\textbf{Downstream Datasets}
To evaluate the downstream performance of our model, we conduct linear probe and zero-shot classifications, as introduced in Sec. \ref{sec:pre_clip}, on 10 widely used datasets \citep{radford2021learning,li2021supervision,yang2023robust} listed in Appendix, Table \ref{tab:downstream_datasets}. 

\textbf{Adversarial Attacks}
To evaluate the effectiveness of our defense, we consider five different attack baselines: targeted data poisoning attacks (TDPA), backdoor attacks (BA) with visible triggers like BadNet, with invisible triggers like Blended and WaNet, and label consistent backdoor attacks. Examples of different backdoor patterns are presented in Appendix Fig. \ref{fig:attack} \citep{carlini2021poisoning,gu2017badnets,nguyen2021wanet,chen2017targeted,turner2019label}, 
For TDPAs, we randomly select 16 different images from the CC3M validation set as our target images.  
For each target image, we choose a random class from the ImageNet1K dataset \citep{deng2009imagenet}, and construct an adversarial caption set related to the label as discussed in Sec. \ref{sec:pre_attack}. 
We set the poison rate for all datasets as 0.05\%. 
For BAs, we randomly select images from the CC3M validation data and apply the corresponding backdoor triggers. For each attack, we choose a random class from the ImageNet1K dataset \cite{deng2009imagenet} and construct the adversarial caption set related to the label as discussed in Sec. \ref{sec:pre_attack}. Each backdoored image is paired with a random poisoned caption from the adversarial caption set. Following \cite{bansal2023cleanclip}, we set the backdoor rate for BadNet Attack to 0.05\%, and 
the backdoor rate for other four backdoor attacks to 0.15\% (otherwise they cannot poison CLIP successfully).

\textbf{Defense Baselines}
We consider RoCLIP, the only existing pre-training defense, as our baseline \citep{yang2023robust}. RoCLIP pairs
each image representation with its nearest neighbor caption in a pool of random caption representations. 
We measure the effectiveness of attacks using attack success rate (ASR). For TDPA, ASR is measured as the fraction of target images that are classified as the adversarial label. For BA, ASR is measured as the fraction of test images containing the backdoor triggers that are classified as the adversarial label. 

\vspace{-1mm}
\subsection{\alg Defends CLIP \& Preserves Performance} \label{sec:res}\vspace{-1mm}
Here, we evaluate the performance of \alg against TDPA and BAs. We compare \alg with CLIP and RoCLIP, based on both ASR and downstream performance. Table \ref{tab:defense} shows that adversarial attacks are highly effective against CLIP, with ASRs over 60\% for TDPA on all datasets and above 90\% for some BAs. This highlights the significant challenge of ensuring CLIP robustness. \alg effectively reduces the ASR to nearly 0\% across all datasets for both TDPA and BAs. Even in TDPA where only a few images are targeted, \alg's defense is strong, with very few successful attacks. 
We see that while RoCLIP and \alg can both defend the model relatively well, RoCLIP is less consistent than \alg.
Notably, RoCLIP's ASR on TDPA in the VG dataset is 37.5\% higher than \alg’s, and on Blended is 4.6\% higher. 
Importantly, Table \ref{tab: Classification Utility} shows that while \alg maintains a comparable performance to CLIP, RoCLIP significantly harms the overall performance by nearly 10\% on both zero-shot and linear probe. \looseness=-1
\subsection{\alg Ablation Study and Sensitivity Analysis} \label{sec:abl_warmup}
\alg warms up the model 
by applying 5 epochs unimodal CL followed by applying CLIP loss to all examples once with $lr_\text{low} = 5e^{-6}$. Here, we illustrate the necessity of each of these components. We conduct our experiments on various datasets against TDPA with a poison rate of 0.05\%.\looseness=-1

\begin{table}[htbp]
\centering
\caption{Effect of \# of CL warmup epochs and \# of times CLIP loss is applied to examples with lower learning rate.}
\vspace{-2mm}
\begin{small}
\begin{tabular}{ccc}
\toprule
\textbf{\# CL epochs} & \textbf{\# CLIP epochs} & \textbf{Poison Rate in $\mathcal{D}_\text{safe}$} \\
\midrule
\textbf{5} & \textbf{1} &  \textbf{0.09\%} \\
1 & 1 & 1\% \\
10 & 1 & 0.03\% \\
5 & 0 & 12.28\% \\
5 & 2 & 6.8\% \\
\bottomrule
\end{tabular}
\label{tab:unimodal_clip}
\end{small}
\end{table}



\textbf{Impact of Unimodal CL Warmup}
Table \ref{tab:unimodal_clip} shows the proportion of poisoned data remained in the safe set after warmup. Rows 1 to 3 indicates that, increasing unimodal training epochs significantly lowers the poison rate. Specifically, we observe a 0.91\% drop in the poison rate when increasing the number of unimodal CL epochs from 1 to 5. However, extending the warmup duration beyond 5 epochs results in diminishing returns, i.e., 
10 epochs of unimodal training only marginally reduces the poison rate in the safe set from 0.09\% to 0.03\%.
In our experiments, we consistently apply 5 epochs of unimodal CL warmup across various datasets and attack types. As shown in Table \ref{tab:defense}, this approach yields robust defense across different scenarios, confirming its broad effectiveness.

\textbf{Impact of CLIP loss with Low Learning Rate} 
Next, we conduct an ablation study on the number of times CLIP loss is applied with low learning rate to all examples. As shown in Table \ref{tab:unimodal_clip}, rows 1 and 4, in the absence of any CLIP warmup, 12.28\% of the poisoned pairs remain in the safe set. This occurs because, without any CLIP training, the image representations do not correlate well with the caption representations. On the other hand, it is critical to avoid extensive training with the CLIP loss on the full dataset before filtering out the poisoned pairs. As shown in row 5, applying even one additional epoch of CLIP training with a low learning rate of $5e^{-6}$ introduces 6.8\% more poisoned pairs in the safe set.
\begin{table}[h]
\centering
\caption{Attack success rates of \alg against TDPA on various datasets, with differing values of low learning rates when applying CLIP loss to separate safe and risky sets. \looseness=-1}
\label{tab:slow_lr}
\vspace{-2mm}
\begin{small}
\begin{tabular}{@{}llll@{}}
\toprule
\textbf{$lr_\text{low}$} & \textbf{$5e^{-6}$} & \textbf{$1e^{-5}$} & \textbf{$5e^{-5}$} \\ 
\midrule
CC1M             & 0\%        & 0\%        & 0\%        \\
COCO             & 0\%           & 0\%           & 0\%           \\
VG               & 0\%           & 0\%           & 0\%           \\
\bottomrule
\end{tabular}
\end{small}
\end{table}
Next, we explored the learning rate's sensitivity during the slow-paced CLIP warmup, with results shown in Table \ref{tab:slow_lr}. We examined a range of low learning rates from $5e^{-6}$ to $5e^{-5}$ across various datasets and found consistent strong defense against TDPA. This indicates that \alg is not sensitive to $lr_\text{low}$ 
and does not require precise tuning.

\textbf{Impact of unimodal CL during Pre-training}
\label{sec:mixed_train} \alg applies unimodal CL to the risky data during pre-training. 
Table \ref{tab:mt_nn} shows that applying unimodal CL to the risky set is essential, to prevent poisons from getting into the safe set and poisoning the model. Otherwise, the ASR significantly increases at the end of training across all datasets. \looseness=-1

\textbf{Impact of Nearest Neighbor}
We also conducted experiments on the impact of nearest-neighbors on \alg defense. Table \ref{tab:mt_nn} shows that, without the NN pool, the ASR significantly increases at the end of training across all datasets. \looseness=-1

\textbf{Impact of GMM Threshold} 
Using a lower GMM threshold allows \alg to train with more data, but it significantly increases the risk of the model being poisoned. Table \ref{tab:threshold_data} demonstrates that across all datasets, a lower threshold leads to a significant increase in poison rates. For instance, reducing the threshold from 0.9 to 0.5 results in 8 times more poisoned pairs in COCO, 17 times more in CC1M, and a 0.3 increase in poison rate for VG. Conversely, a higher threshold, while reducing the poison rate, leads to substantial performance losses for \alg due to reduced training data. For example, a 0.05 threshold increase results in CC1M being trained on half the data and COCO on over ten times less data. These findings highlight the importance of an optimal threshold for \alg's effectiveness. Through extensive experimentation, we determined that a threshold of $t=0.9$ works well for all datasets, by making a balance between large training data and maintaining a low poison rate.\looseness=-1



\begin{table}[htbp]
\centering
\caption{Impact of applying unimodal CL to risky set, or not using Nearest Neighbor (NN) with CL.}
\label{tab:mt_nn}
\vspace{-2mm}
\small
\begin{tabular}{@{}lccc@{}}
\toprule
\textbf{Dataset} & \textbf{\alg} &{ASR (No CL)} & {ASR (No NN w. CL)} \\
\midrule
CC1M    & 0\% & 12.5\%     & 12.5\%          \\
COCO    &0\% & 12.5\%     & 25.0\%    \\
VG      &0\% & 12.5\%     & 25.0\%    \\
\bottomrule
\end{tabular}
\end{table}

\begin{table}[ht]
\centering
\caption{
Total poison ratio of the safe set after filtering with different GMM thresholds. The ratio of data in safe set is shown in parentheses. The initial poison rate is 0.05\%.}
\label{tab:threshold_data}
\vspace{-2mm}
\begin{small}
\setlength{\tabcolsep}{2pt}
\begin{tabular}{@{}l|c|c|c|c@{}}
\toprule
Dset &$t=0.95$ & $\textbf{t=0.9}$  & $t=0.7$  & $t=0.5$ \\ 
\midrule
CC1M  & $0$ (5.8) &\textbf{$3.75e^{-4}$ (11.4)}  & $3.13e^{-3}$ (29.0) & $6.25e^{-3}$ (45.2) \\
COCO  & $0$ (0.7)&\textbf{$2.5e^{-3}$ (10.3)} & $1.13e^{-2}$ (42.3) & $2.13e^{-2}$ (62.7)  \\
VG     &$0$ (0.0) &\textbf{$0$~~~~~~~~~~ (7.2)} & $7.5e^{-3}$ (45.3) & $1.88e^{-2}$ (65.6) \\
\bottomrule
\end{tabular}
\end{small}
\end{table}
\textbf{\alg's Usage of Pre-training Data}
\alg\ applies the CLIP loss only to the data in the safe set to protect the model. Thus, it only pre-trains CLIP on a fraction of data.
Nevertheless, \alg\ can benefit from more data with extended training. To confirm this, we extend our experiment on CC1M to 64 epochs and attack the models with targeted data poisoning (TDPA) and BadNet backdoor attacks. The results are shown in Table \ref{tab:data_eff}. By the end of training, 80\% of the data is included in the safe set. \alg achieves much higher zero-shot and linear probe accuracy on CIFAR-10, CIFAR100, and ImageNet1K, confirming that \alg\ can 
effectively utilize more data. 
Notably, longer training with \alg\ does not introduce more poisoned pairs in the safe set, and the ASR remains unchanged. \looseness=-1
\begin{table}[ht]
\centering
\caption{Extended training for 64 epochs effectively improve the data usage of \alg and its performance. }
\label{tab:data_eff}
\vspace{-2mm}
\begin{small}
\setlength{\tabcolsep}{2pt}
\begin{tabular}{llccc|cc}
\toprule
Method & Task & C10 & C100 & I1K & \textit{TDPA} & \textit{BadNet} \\
\midrule
\multirow{2}{*}{\alg} & 0-shot & 39.7 & 10.41 & 9.87 &\multirow{2}{*}{0\%} & \multirow{2}{*}{0\%}\\
 & lin-prb & 71.9 & 47.32 & 24.53 & &\\
\midrule
\multirow{2}{*}{\alg-64} & 0-shot & \textbf{43.1} & \textbf{14.4} & \textbf{12.6} &\multirow{2}{*}{0\%} & \multirow{2}{*}{0\%} \\
 & lin-prb & \textbf{75.0} & \textbf{50.6} & \textbf{28.7} \\
\midrule
\multirow{2}{*}{CLIP} & 0-shot & 34.9 & 7.3 & 9.6  &\multirow{2}{*}{93.8\%} & \multirow{2}{*}{100\%} \\
 & lin-prb & 70.5 & 45.8 & 22.2  \\
\bottomrule
\end{tabular}
\end{small}
\end{table}

\textbf{\alg's Complexity and Overhead}
Next, we measure \alg's average overhead per epoch on all datasets, relative to standard CLIP 
pre-training. For reference, each CLIP epoch is considered equivalent to a value of 1. Table \ref{tab:structure_time} shows that every \alg's unimodal CL warm-up epoch 
and applying the CLIP loss with small learning rate 
take a similar amount of time to a CLIP pre-training epoch. 
Separating the safe set from the risky set with GMM takes approximately 0.35 times the duration of a CLIP epoch, but is required only once during the entire training. Updating safe and risky sets takes 0.1 of a CLIP epoch time. While RoCLIP is slightly more efficient than \alg, it is important to highlight the significant difference in their downstream performances, as demonstrated in Table \ref{tab: Classification Utility}. We include \alg's efficient implementations and its time complexity compared to popular defense methods in supervised learning settings in Sec. \ref{sec:overhead}. Compared to such defenses, \alg\ is orders of magnitude more efficient.

\begin{table}[ht]
\centering
\caption{Time complexity of \alg\ relative to CLIP.}
\label{tab:structure_time}
\vspace{-2mm}
\begin{small}
\setlength{\tabcolsep}{8pt}
\begin{tabular}{ll}
\toprule
\textbf{Structure} & \textbf{Time} \\
\midrule
\alg: Unimodal CL epoch & 1 \\
\alg: Pre-training epoch (CLIP+CL) & 1 \\
\alg: GMM (required only once) & 0.35 \\
\alg: Updating Safe \& Risky sets & 0.1 \\
CLIP epoch & 1 \\
RoCLIP epoch & 1.06 \\
\bottomrule
\end{tabular}
\end{small}
\end{table}

\textbf{Effectiveness of \alg on Different Data Scales} We conduct an ablation study to examine the effectiveness of \alg on different subsets of CC3M during warm-up epochs. We study two factors: the fraction of examples in the safe set, which reflects the model's final performance, and the fraction of poisoned examples per attack that remained in the safe set, which indicates the risk of model poisoning during pre-training. We considered TDPA and BadNet with a poison rate of 0.05\%. Table \ref{tab:safe_set} demonstrate that \alg can consistently reduce the poison rate across datasets of different sizes. Larger datasets allow \alg to find a larger safe set after warm-up, improving its data usage. Notably, the poison rate within the safe set decreases significantly as dataset size increases, particularly from 100K to 1M. This indicates that \alg can effectively protect CLIP pre-training on large datasets.

\subsection{\alg is Robust against Adaptive Attacks}
Next, we discuss two potential adaptive attacks against \alg, and show that they cannot affect \alg. 

\textbf{Attacks against Unimodal CL}
Unimodal CL training during both the warmup and pre-training phases is crucial for \alg to separate poisoned data from clean data. However, this makes \alg susceptible to adversarial attacks targeting unimodal CL, e.g. those proposed in \cite{kim2020adversarial}, where backdoor triggers are patched onto unlabeled training images of a chosen target category. If such images are included in the risky set and are trained on with unimodal CL, they could risk backdooring \alg during linear probe evaluation. Next, we show that \alg remains robust against these attacks. 
For inclusion in the CLIP pre-training data, backdoored images need to be paired with captions. If paired with captions from another category, they make a version of targeted data poisoning attack that we have already studied in our paper. Therefore, we assume the target images are paired with correct category captions. 
Given the low backdoor rate and the dissimilarity of backdoored images to their category, such images do not align closely with the adversarial text category after applying the CLIP loss with low learning rate and end up in the risky set. 
Using the NN pool and data augmentation in \alg's unimodal CL effectively counters backdoor attacks. Given the low backdoor rate, these methods prevent clustering of such images in the representation space. As the backdoored images do not end up in the safe set and do not cluster tightly in the image representation space, they cannot poison the model (zero-shot or linear-probe evaluation). 
To confirm this, we conduct experiments on CC1M dataset with increased backdoor rates (up to 0.15\%). Post-warmup, 94.5\% of backdoored images were in the risky set, yet \alg maintained a 0\% ASR in linear-probe classification, underscoring its resilience to these adaptive attacks.\looseness=-1

\textbf{Attacks against Semi-supervised Learning}
Adversarial attacks on semi-supervised learning models, such as those described in \cite{carlini2021poisoningb}, pose another potential threat. In these attacks, adversary generates unlabeled images that interpolate between a labeled image $\x_i$ from the target category and an unlabeled image $\x_j$. The goal is to cause $\x_j$ to be misclassified as the target category. 
To be added to CLIP pre-training data, poisoned images required to be paired with captions. In particular, to be misclassified as target, adversarial captions should belong to the target category. Among the interpolated images, the ones that are more similar to $\x_i$ do not pose any risk of poisoning for the model. The ones that are more similar to $\x_j$ but are paired with target-related captions act as a weaker targeted data poisoning attack on CLIP, which we have studied in our paper. 
\alg effectively identifies such examples as risky, and pre-trains CLIP robustly against such attacks.\looseness=-1

\vspace{-1mm}
\subsection{\alg\ is Robust against Stronger Attacks}
\textbf{Attacks with Higher Poison Rate} Our experiments already confirm the effectiveness of \alg for poison ratio up to 0.05\%. Here, we explore a higher poison ratio of 0.1\% and 0.5\% using the same hyperparameter setting on MSCOCO and CC1M. 
Table \ref{tab:high_poison_rate} shows that with a higher poison ratio, \alg can still defend the attack successfully.\looseness=-1

\begin{table}[ht]
\centering
\caption{\alg defends attacks with higher poison rate}
\label{tab:high_poison_rate}
\vspace{-2mm}
\begin{small}
\setlength{\tabcolsep}{4pt} 
\begin{tabular}{l|cc|ccc}
\toprule
 & \multicolumn{2}{c|}{\textbf{TDPA}} & \multicolumn{3}{c}{\textbf{BadNet}} \\
\textbf{Poison Rate} & \textbf{0.05\%} & \textbf{0.1\%} & \textbf{0.05\%} & \textbf{0.1\%} & \textbf{0.5\%} \\
\midrule
MSCOCO & 0\% & 0\% & 0\% & 0\% & 0\% \\
CC1M & 0\% & 6.25\% & 0\% & 0\% & - \\
\bottomrule
\end{tabular}
\end{small}
\end{table}

\textbf{Multi-trigger backdoor attacks} We also study \alg's effectiveness against multi-trigger backdoor attacks \citep{li2024multi} on MSCOCO. We considered two backdoor strategies:
(1) \textit{Hybrid-Trigger Backdoor Attack (HTBA),} where for every backdoored image, 
multiple (distinct) triggers are patched onto the image; 
(2) \textit{Parallel Backdoor Attack (PBA),} where multiple distinct subsets of images with different backdoor triggers are injected into the pre-training dataset. Each subset may correspond to either the same target class (All2One),
or different target classes (All2All).
In parallel and hybrid-trigger backdoor attacks, we consider the same backdoor triggers from the main experiments, namely BadNet, WaNet, and Blended.
For All2One attacks and HTBA, we consider a random target category and a total poison rate of 0.05\%. For All2All attacks,  
we select three random categories (one for each backdoor trigger) as the target classes, with a poison rate of 0.05\% for each.
Table \ref{tab:multi_trigger} shows that \alg can successfully defend all the attacks and reduce the ASR to 0\%.\looseness=-1

\begin{table}[ht]
\centering
\caption{\alg against Multi-trigger backdoor attacks.}
\label{tab:multi_trigger}
\vspace{-2mm}
\begin{small}
\setlength{\tabcolsep}{8pt} 
\begin{tabular}{lcc}
\toprule
\textbf{Strategy} & \textbf{Attack Success Rate} \\
\midrule
HTBA& 0\% \\
PBA: All2One 
& 0\% \\
PBA: All2All 
& 0\% \\
\bottomrule
\end{tabular}
\end{small}
\end{table}

\vspace{-3mm}
\section{Conclusion}
We proposed \alg, an effective method for safely pre-train CLIP against targeted data poisoning and backdoor attacks. 
Using unimodal CL warmup and CLIP warmup with low learning rate, \alg filters majority of the poisons before pre-training and defends the model during pre-training by applying the CLIP loss to pairs with high similarity and applying unimodal CL to rest of the examples.
We showed that \alg 
lowers the success rate of targeted data poisoning attacks from 93.75\% to 0\% and that of various backdoor attacks from as high as 100\% to 0\%, without adversely affecting CLIP's performance on various datasets. \looseness=-1

\section*{Impact Statement}
This paper presents work whose goal is to advance the field of Machine Learning. There are many potential societal consequences of our work, none which we feel must be specifically highlighted here.

\section*{Acknowledgments} This research was supported by the National Science Foundation CAREER Award 2146492 and Cisco Systems.
\bibliography{main}
\bibliographystyle{icml2024}
\clearpage
\newpage
\section{Appendix}
\subsection{Benchmark Datasets}
\label{sec:dataset}
\textbf{Pretrain dataset}
MSCOCO: MSCOCO \citep{lin2014microsoft} is a large-scale dataset for object detection, segmentation, and captioning. It features 80 object categories, with each image accompanied by 5 captions. For our analysis, we randomly select one caption per image. The total dataset size is 80K images.

Visual Genome: Visual Genome \citep{krishna2017visual} is a comprehensive dataset for region captions. It comprises 10877 images and 5.4 million region descriptions. For each image, we randomly select 5 region descriptions and merge them into one caption.

Conceptual Captions: Conceptual Captions \citep{sharma2018conceptual} is a vast, web-scale image captioning dataset that encompasses a wide variety of image styles and caption formats.

\textbf{downstream dataset}
To evaluate the downstream performance of our model, we conduct linear probe and zero-shot classifications, as introduced in Sec. \ref{sec:pre_clip}, on 10 widely used datasets \citep{radford2021learning,li2021supervision,yang2023robust} listed in Table \ref{tab:downstream_datasets}.

\begin{table}[h!]
\caption{Details of downstream datasets.}
\label{tab:downstream_datasets}
\vspace{-5mm}
\begin{center}
\begin{small}
\begin{tabular}{llrr}
\toprule
\textbf{Dataset} & \textbf{Classes} & \textbf{Train Size} & \textbf{Test Size} \\
\midrule
CIFAR10 & 10 & 50,000 & 10,000\\
CIFAR100 & 100 & 50,000 & 10,000\\
Food-101 & 101  & 75,750 & 25,250\\
DTD & 47 & 3,760 & 1,880 \\ 
FGVC Aircraft & 100 & 6,667 & 3,333 \\
Flowers-102 & 102 &  2,040 & 6,149\\
Caltech-101 & 102 & 3,060 & 6,085\\
OxfordIIITPet & 37  & 3,680 & 3,669\\
Stanford Cars & 196 & 8,144 & 8,041\\
ImageNet1K & 1000  & 50,000 & 50,000\\
\bottomrule
\end{tabular}
\end{small}
\end{center}
\vspace{-4mm}
\end{table}

\begin{figure}[ht]
    \centering
    \begin{subfigure}{.32\columnwidth}
        \centering
        \includegraphics[width=\linewidth]{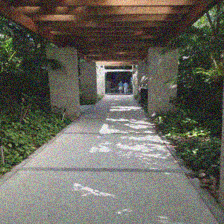}
        \caption{Blended}
    \end{subfigure}
    \hfill
    \begin{subfigure}{.32\columnwidth}
        \centering
        \includegraphics[width=\linewidth]{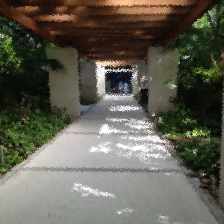}
        \caption{WaNet}
    \end{subfigure}
    \hfill
    \begin{subfigure}{.32\columnwidth}
        \centering
        \includegraphics[width=\linewidth]{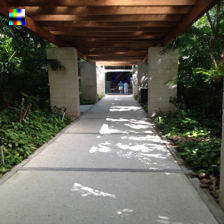}
        \caption{BadNets}
    \end{subfigure}
    \caption{Backdoor attacks used in our evaluations.}
    \label{fig:attack}
    
\end{figure}

\begin{table}[ht]
\centering
\caption{Defense of \alg on datasets of different sizes. Safe Set \% indicates the fraction of examples in the safe set after warm up, and Safe Set Poison Rate indicates the fraction of poisoned examples per attack that remained in the safe set}
\label{tab:safe_set}
\vspace{-2mm}
\begin{small}
\setlength{\tabcolsep}{10pt}
\begin{tabular}{lcc}
\hline
\textbf{Dataset} & \textbf{Safe Set \%} & \textbf{Safe Set Poison Rate} \\
\hline
CC3M & 17.79\% & 0.000606\% \\
CC1M & 11.38\% & 0.000375\% \\
CC100K & 6.25\% & 0.00176\% \\
Unfiltered & 100\% & 0.05\% \\
\hline
\end{tabular}
\end{small}
\end{table}

\begin{table}[ht]
\centering
\caption{\alg with different model structures}
\label{tab:model_poison_rate}
\vspace{-2mm}
\begin{small}
\setlength{\tabcolsep}{10pt}
\begin{tabular}{lcc}
\toprule
\textbf{Model Structure} & \textbf{TDPA} & \textbf{BadNet} \\
\midrule
ResNet50 & 0\% & 0\% \\
ViT-B/32 & 0\% & 0\% \\
\bottomrule
\end{tabular}
\end{small}
\end{table}

\begin{table}[h]
\centering
\caption{{Hyperparameters of our experiments}}\label{tab:tuning}
\begin{tabular}{l|l|l|l}
\toprule
\textbf{Dataset} & $lr_\text{low}$ & \textbf{lr} & \textbf{Batch Size} \\ 
\midrule
CC3M & 5e-6 & 5e-4 & 512 \\
CC1M & 5e-6 & 5e-4 & 512 \\
COCO & 5e-6 & 5e-4 & 256 \\
VG   & 5e-6 & 5e-4 & 256 \\
\bottomrule
\end{tabular}
\end{table}

\begin{algorithm}[!t]
   \caption{\alg} \label{alg:alg}
   \begin{small} 
\begin{algorithmic}
   \STATE \textbf{Input:} Image encoder $f_I$, text encoder $f_T$, image pool $P_I$, text pool $P_T$, unimodal warmup epochs $r$, training epochs $T$, GMM threshold $t$, increment ratio $s$, small learning rate $l_{low}$
   \STATE \textbf{Data:} Dataset of image-caption pairs $\mathcal{D}=\{(x_i^\I, x_i^\T)\}_{i=1}^n$, $\X_I=\{x_i^\I\}_{i=1}^n, \X_T=\{x_i^\T\}_{i=1}^n$
   \FOR{$epoch = 1$ \textbf{to} $r$}
       \STATE Train $f_I$ with $\Loss_{\text{unimodal\_NN}}(\X_I, P_I)$ in Eq \ref{eq:unimodal_nnclip}
       \STATE Train $f_T$ with $\Loss_{\text{unimodal\_NN}}(\X_T, P_T)$ in Eq. \ref{eq:unimodal_nnclip}
   \ENDFOR
   \STATE Update $f_I, f_T$ by training on $\mathcal{D}$ with $\Loss_{CLIP}$ in Eq. \ref{eq:clip} using $l_{low}$
   \FOR{$epoch = r$ \textbf{to} $T$}
    \STATE $p_{\text{clean}} \leftarrow \text{GMM}\left(f_I(x_i^{\I}), f_I(x_i^{\T})\right)$
    \IF{$epoch = r$}
        \STATE $\D_{\text{safe}} \leftarrow$ all data where $p_{\text{clean}} > t$, filtering out a safe set of training ratio $m\%$
    \ELSE
        \STATE Sort $p_{\text{clean}}$ in a decreasing manner 
        \STATE $\D_{\text{safe}} \leftarrow$ top $m\%$ of data 
    \ENDIF
    \STATE $\D_{\text{risky}} \leftarrow \D \setminus \D_{\text{safe}}$
    \STATE $\D_{\text{safe\_aug}} \leftarrow$ augmented examples in $\D_{\text{safe}}$
    \STATE Train $f_I, f_T$ with $\mathcal{L}_{\text{\alg}}(\D) = \mathcal{L}_{\text{unimodal\_NN}}(\D_{\text{risky}}) + \mathcal{L}_{\text{CLIP}}(\D_{\text{safe\_aug}})$
    \STATE $m \leftarrow m + s$
\ENDFOR
\end{algorithmic}
\end{small}
\end{algorithm}

\subsection{\alg's Complexity and Overhead}
\label{sec:overhead} We note that although \alg introduces additional overhead, different steps can be implemented efficiently without compromising the model’s performance or defense capabilities, as discussed below.

\textbf{Number of CL Epochs}: As shown in Table \ref{tab:unimodal_clip}, the effect of more unimodal CL epochs diminishes after 1 epoch, and even 1 epoch of unimodal CL is enough to filter most of the poisoned examples. After 5 epochs, there is no benefit for unimodal CL on any dataset. We observed this trend on various datasets of different sizes and distributions and do not expect \alg to require more than 5 unimodal CL epochs on any dataset.

\textbf{Data Partitioning to Safe and Risky Sets}: As demonstrated in Figure \ref{fig:gmm}, following the warm-up phase, the majority of the poisoned data pairs have a low cosine similarity. Note that at every epoch, \alg only incorporates an additional 1\% of data from the risky set. Therefore, to update the safe and risky sets at the beginning of every epoch, \alg only needs to re-evaluate a portion of the data with high cosine similarity from the previous epoch, rather than the entire dataset. This approach significantly reduces the overhead associated with the method. In our experiments, we only re-evaluate the cosine similarities of the top 30\% of the dataset from the previous epoch, which reduced the overhead to 0.1 CLIP epoch time. On all our training datasets, this efficient implementation obtains a safe set with a similar average poison rate compared to the safe set obtained via full data evaluation.

Compared to popular defense methods in the supervised setting, \alg has a small overhead. With 5 unimodal CL warm-up epochs, \alg increases the total training time by about 29.84\%. In contrast, supervised defenses \citep{peri2020deep,geiping2021doesn,chen2018detecting,tran2018spectral} increase training time by up to 866.67\%. We see that the additional computational overhead of \alg is relatively low compared to supervised defense methods.

\begin{table}[ht]
\centering
\caption{Training time of \alg compared to supervised defense methods. We measure the increased training time of different methods compared to their regular training time without defenses.}
\label{tab:training_time}
\vspace{-2mm}
\begin{small}
\setlength{\tabcolsep}{8pt}
\begin{tabular}{ll}
\hline
\textbf{Method} & \textbf{Increased Time} \\
\hline
DeepKNN & 866.67\% \\
Spectral Signatures & 166.67\% \\
Activation Clustering & 106.67\% \\
Adv. Poisoning & 653.33\% \\
\alg (5 CL epoch) & 29.84\% \\
\alg (1 CL epoch) & 17.34\% \\
\hline
\end{tabular}
\end{small}
\end{table}

\subsection{Hyperparameter Tuning}
We include the hyperparameter settings of our experiments in Table \ref{tab:tuning}. There are few key hyperparameters for tuning:

\textbf{Number of CL epochs}: As shown in Table \ref{tab:unimodal_clip}, the effect of more unimodal CL epochs diminishes after 1 epoch, and even 1 epoch of unimodal CL is enough to filter most of the poisoned examples. After 5 epochs, there is no benefit for unimodal CL on any of the datasets. We observed this trend on various datasets of different size and distribution and we do not expect \alg to require more than 5 in-modal CL epochs on any dataset.

\textbf{Small learning rate}: For the small learning rate of training 1 epoch with CLIP loss, we showed in table \ref{tab:slow_lr} that \alg is not sensitive to the choice of the small learning rate and about 0.01x the original learning rate works well on various datasets with different distributions and sizes.

\textbf{\alg with different model architecture} CLIP has two variations in its vision model, ResNet, which we used in our original experiments, and ViT, which we report here. We attack the model with TDPA and BadNet backdoor with a poison rate of 0.05\%, consistent with the setting in the paper. As shown in Table \ref{tab:model_poison_rate}, in both architectures, \alg can defend the model with the same hyperparameter setting.

\textbf{Limitation.}
\alg warms up the model with in-modality CL followed by 1 CLIP epoch with small learning rate to distinguish the clean and poisoned pairs. However, if the number of injected poisons are too high, \alg may not be able to distinguish the poisoned pairs from the clean pairs. From our experiments, we were not able to effectively distinguish the majority of poisoned pairs after warmup, when poison rate is as high as 0.5\%. If a small clean dataset of image-caption pairs is available, \alg can leverage that to defend a much higher poison rate.

\end{document}